\documentclass[letterpaper, 10pt, conference]{ieeeconf}  % Comment this line out if you need a4paper

\IEEEoverridecommandlockouts                              % This command is only needed if 
\usepackage{amsmath} % assumes amsmath package installed
\usepackage{amssymb}  % assumes amsmath package installed
\makeatletter
\let\NAT@parse\undefined
\makeatother
\usepackage[numbers]{natbib}
\let\labelindent\relax
\usepackage{enumitem}
\usepackage{tikz}
\usepackage{pgfplots}
\usepackage{pgfplotstable}
\usepgfplotslibrary{groupplots}
\usepackage{subcaption}
\usepackage{multirow}
\usepackage{placeins}

\definecolor{mpi-green}{RGB}{0,117,103}
\definecolor{mpi-grey}{RGB}{211,211,205}
\definecolor{mpi-lblue}{RGB}{58,185,198}
\definecolor{mpi-blue}{RGB}{45,147,207}
\definecolor{mpi-red}{RGB}{204,75,99}
\definecolor{mpi-lgreen}{RGB}{75,168,79}
\definecolor{mpi-purple}{RGB}{114,122,178}
\definecolor{mpi-orange}{RGB}{247,179,94}
\definecolor{plot-orange}{RGB}{229,133,11}
\definecolor{plot-grey}{RGB}{182,183,175}

\usepackage{comment}

\title{\LARGE \bf Accurate Vision-based Manipulation through Contact Reasoning}

\author{Alina Kloss$^{1}$, Maria Bauza$^{2}$, Jiajun Wu$^{2,3}$, Joshua B. Tenenbaum$^{2}$, Alberto Rodriguez$^{2}$ and Jeannette Bohg$^{1,3}$
\thanks{$^{1}$ Max Planck Institute for Intelligent Systems, akloss@tue.mpg.de}
\thanks{$^{2}$ Massachusetts Institute of Technology,
\{bauza, jbt, albertor\}@mit.edu}
\thanks{$^{3}$ Stanford University, jiajunwu@cs.stanford.edu, bohg@stanford.edu}
}

\begin{document}
\maketitle
\thispagestyle{empty}
\pagestyle{empty}

%%%%%%%%%%%%%%%%%%%%%%%%%%%%%%%%%%%%%%%%%%%%%%%%%%%%%%%%%%%%%%%%%%%%%%%%%%%%%%%%
\begin{abstract}
Planning contact interactions is one of the core challenges of many robotic tasks. 
Optimizing contact locations while taking dynamics into account is computationally costly and,
in environments that are only partially observable, executing contact-based tasks often suffers 
from low accuracy. We present an approach that addresses these two challenges for the 
problem of vision-based manipulation. First, we propose to disentangle contact from motion 
optimization. Thereby, we improve planning efficiency by focusing computation on promising 
contact locations. Second, we use a hybrid approach for perception and state estimation that 
combines neural networks with a physically meaningful state representation. In simulation and 
real-world experiments on the task of planar pushing, we show that our method is more efficient 
and achieves a higher manipulation accuracy than previous vision-based approaches.
\end{abstract}

%%%%%%%%%%%%%%%%%%%%%%%%%%%%%%%%%%%%%%%%%%%%%%%%%%%%%%%%%%%%%%%%%%%%%%%%%%%%%%%%
\section{Introduction}

In many robotics applications that involve manipulation or legged locomotion, planning 
contact interactions is one of the core challenges. 
The main problems are the computational cost of optimization and the uncertainty induced by 
imperfect perception. 
Current approaches roughly fall into two categories that align with these problems. 
The first focuses on reducing the computational cost of motion optimization especially for 
long sequences and complex dynamics. 
Such approaches typically make the strong assumption of a fully observable state and 
prior knowledge of the robot and environment, which are rarely fulfilled in practice. 
The second category focuses on including perception and being robust to the resulting uncertainty.
Approaches in this category are typically learning-based and provide a larger level of 
generalization to variations of the environment, such as unknown objects. However, this often 
comes at the cost of accuracy. We propose an approach that addresses both main challenges 
in planning contacts, imperfect visual perception and the computational
complexity of the task. We show that by combining  learning-based perception with an 
explicit state representation, we can achieve accuracy and generalization, while disentangling
contact and motion optimization allows for efficient planning.

\begin{figure}[tb!]
 \centering
 \begin{subfigure}{0.7\linewidth}
  \centering
 \includegraphics[height=2.2cm ]{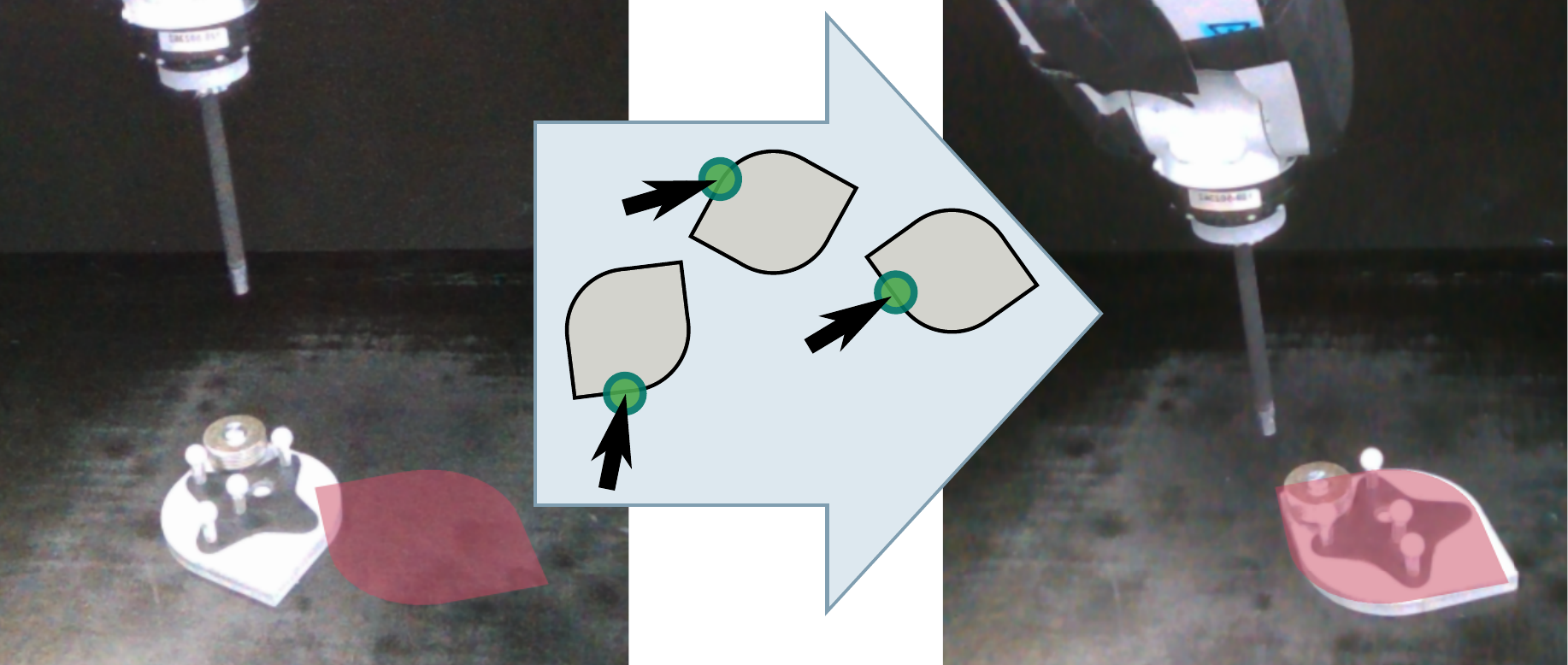}
 \caption{Planar Pushing\label{fig:teaser}}
 \end{subfigure}
 \begin{subfigure}{0.25\linewidth}
  \centering
 \includegraphics[height=2.2cm ]{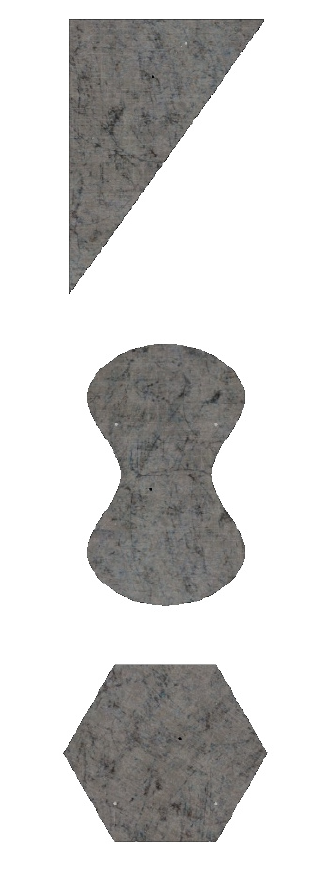}
 \caption{Test objects\label{fig:objects}}
 \end{subfigure}
 \caption{(a) To push an object to a desired pose (red), a robot has to 
 reason over {\em where} (green contact points) and {\em how} (black arrows) to push. (b) The test objects 
 triangle, butter and hexagon.}
\end{figure}

As a case study, we use quasi-static planar pushing with a point contact. This non-prehensile 
manipulation primitive 
can be represented by a simple, low-dimensional state, but has surprisingly complex dynamics,
making it difficult to control.
Prior work can be split according to the aforementioned two main challenges. Approaches based on 
analytical models and a physically meaningful state representation often achieve high accuracy, but assume 
full observability and known object shape \cite{multistep, coarse-fine, reactive, in-hand, gp-model}. 
Learning-based approaches address the perception problem and make fewer assumptions about the  
environment, but are less accurate \cite{push-net, visual-foresight, visual-foresight2, visual-foresight3, poke, contact-learning, r2.2r6.1}. 
Moreover, many of these works do not explicitly reason over where to push, but sample 
random actions \cite{multistep, coarse-fine, push-net, visual-foresight, visual-foresight2, visual-foresight3}.
We argue that explicitly optimizing contact points makes planning more efficient by 
focusing evaluations on promising regions.

We address the problem of pushing an object to a goal pose given RGBD images, illustrated in Figure \ref{fig:teaser}. 
The robot has to decide {\em where} and {\em how} to push the object. We use a learned model
to capture the shape of the object from the visual input and predict a physically meaningful representation
of the object state. This allows us to use filtering to estimate latent 
variables like the centre of mass of the object online and increase the accuracy of prediction.

Each point on the object outline is densely annotated with approximate predictions of the object motion it affords. 
This allows sampling promising candidates of {\em where} to push the object given a desired target object pose. 
At these contacts, we then optimize {\em how} to push.
For predicting the possible object motion, we compare
an approach based on a physical model to a model-free, learned one.
The learned model makes fewer assumptions and shows advantages in cases that are not well-captured 
by the physical model. 
However, the physics-based model is generally more accurate
and even generalizes to scenarios that violate some of its assumptions.

In summary, we propose a system for planar pushing that:
\begin{itemize}[leftmargin=*]
\item allows for efficient planning by explicitly reasoning about contact-location,
\item improves over model-based approaches by including perception and online estimation of latent object properties,
\item achieves higher accuracy than previous vision-based works by combining learned and analytical elements.
\end{itemize}
We quantitatively evaluate our method in simulation through ablation studies and comparison to state of the art. 
We also demonstrate that it transfers to a real robotic platform.

\section{Related Work}
There is a wide range of research on robotic pushing, from modeling the dynamics, e.g.~\cite{r1.1, r1.5, gp, combining},
to state estimation for pushed objects~\cite{r1.2, r1.4}.
Here, we focus on works that include planning, for a broader review 
see~\cite{survey}. 

\subsection{Efficient Contact Planning under Full Observability}
\citet{reactive} present a real-time controller for tracking a desired trajectory under full observability. 
While the push is locally optimized by a neural network that predicts sticking or sliding, the global contact location is not.
\citet{multistep} present an approach to push an object into a desired pose in multiple steps by combining a global RRT 
planner with a local, sampling based planner. 
\citet{in-hand} reorient a known object in-hand by pushing it against elements
in the workspace. Similar to our work, they use motion-cones to efficiently describe the set of possible object movements. 
\citet{object-based} use a hybrid approach that augments the predictions from a 
physical model with learned residuals to push two disks that are already in contact. The method evaluates
a predefined set of contacts.

Optimizing contacts is also considered in legged locomotion. \citet{footstep} compute a sequence of 
footsteps given a set of obstacle-free regions. For efficiency, the dynamics of the robot are 
not taken into account. To address this issue, \citet{walking} take a similar approach to ours: they train 
an approximate dynamics model over a discrete set of actions that can be used for efficient contact planning 
while taking robot dynamics into account.  

All these approaches assume full observability and known models of dynamics and geometry. 

\subsection{Push Planning under Partial Observability}

\citet{poke} train a network to predict the pushing action required to transform one RGB image 
into another. In contrast, \citet{push-net,visual-foresight, visual-foresight2, visual-foresight3} 
do not directly predict actions but learn a dynamics model for predicting the effect of sampled 
pushes. The input is either a segmentation mask or a full RGB image. Push-Net \citep{push-net} 
samples 1000 actions by pairing pixels inside and outside of the object, while 
\cite{visual-foresight, visual-foresight2, visual-foresight3} sample pusher motions
that are refined iteratively. Neither work reasons over contact locations, whereas our approach directly 
samples promising contact points.
Push-Net~\citep{push-net} also estimates the centre of mass of objects during interaction 
using an LSTM. We rely on an {\em Extended Kalman Filter\/} (EKF) that estimates the physically meaningful 
state representation during interactions.
Similar to our work, \citet{contact-learning} learn a scoring function from histogram features for finding 
suitable contact points. \citet{r2.2r6.1} learn a contact model and a contact-conditioned predictive
model for pushing with a mobile robot.

While making much fewer assumptions, these vision and learning-based methods are less accurate than 
model-based methods. Our method significantly improves on this.

\section{Methods}

\begin{figure*}
\centering
\includegraphics[width=0.85\textwidth]{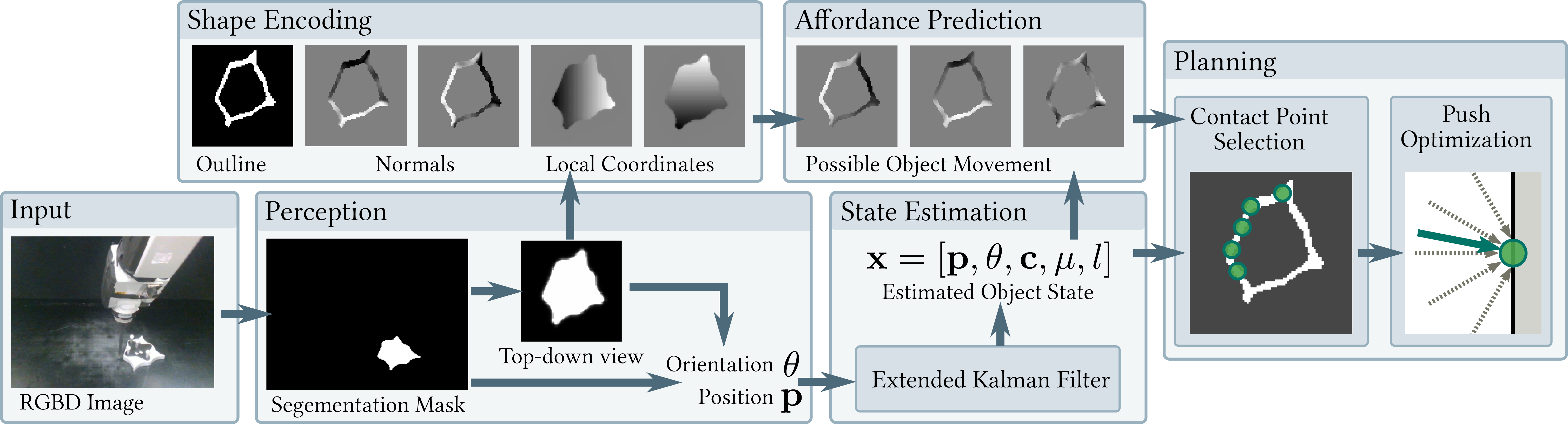}
\caption{Overview: the perception module segments the object and computes its pose.
An EKF estimates the full object state including latent properties like the COM $\mathbf{c}$.
The object shape is encoded by a silhouette, coordinates and normals in a top-down view. It is input to the
affordance prediction module, that approximates the possible object motions at each contact point on the silhouette.
The planning module selects contact point candidates using the predicted affordances and optimizes
the pushing motion there.
}\label{fig:overview}
\end{figure*}

Figure \ref{fig:overview} shows an overview of our system. At each time step,
it receives an RGBD image of the current scene, the last robot action and the 
target object pose as input. In the perception module, we use a {\em Convolutional Neural Network\/} 
(CNN) to segment the object and estimate its position and orientation. Since we do not assume 
prior knowledge of object shape, we extract a representation based on the segmentation map.
Together with the last action, the object pose is input to an EKF that estimates the full object state
including latent properties like centre of mass (COM).

The next module approximates the object motions that can be produced by applying a discrete set of pushes at each point on 
the object silhouette. We refer to the output as \textit{push affordances} of the contact points. 
While this may be considered an abuse of terminology~\citep{OSIURAK2017403}, we use the term for
a clear distinction to other parts in our model.
The affordances are continuously updated as they depend on object properties that are estimated by the EKF,
while the object shape has to be computed only once. 
Finally, the state estimate and affordances are the input to the planning module which selects suitable contact points 
and optimizes the pushing actions beyond the discrete set that is considered in the affordance model.

\subsection{Planar Pushing}

We consider the task of quasi-static planar pushing of a single object using a point contact, where
quasi-static means that the force is enough to move but not to further accelerate the object.
We parametrize a pushing action by the contact point $\mathbf{r}$ and the pushing motion $\mathbf{u}$. 
Pushes are executed at a constant velocity of 20 mm/s.

The dynamics of pushing depend on object shape, friction and pressure distribution of 
the object on the surface. The relation between push force and resulting object motion is 
often modeled using the limit-surface~\cite{howe-cutkosky,goyal}. We use an analytical model by \citet{model}. 
It assumes continuous object-surface contact and an uniform pressure distribution for an ellipsoidal 
approximation of the limit surface parameterized by $l$. The model predicts object translation and rotation 
around the COM given $l$, the push $\mathbf{u}$, the normal $\mathbf{n}$ at the contact point 
and the coefficient of friction between pusher and object $\mu$. 
We use $\mathbf{x} = \begin{pmatrix} \mathbf{p}, \theta, \mathbf{c}, l, \mu \end{pmatrix}$ as object
state, where $\theta$ is the orientation of the object and $\mathbf{c}$ is the position of the COM
relative the object frame origin $\mathbf{p}$.

\subsection{Perception and State Estimation}

We train a CNN to segment the object in each image and
compute its world-frame position from segmentation mask and depth values. A bounding box around the 
segmentation mask is reprojected into a top-down view centered on the object. 
The orientation of the object is computed relative to the first step by comparing stepwise rotations 
of the current top-down projection to the initial one. We also evaluated using a 
neural network for this task but found it to be less reliable.
The output object pose is used as observation for an EKF that estimates the full object state 
$\mathbf{x}$. The filter uses the analytical model~\cite{model} as process model and an identity matrix selecting 
the object pose from $\mathbf{x}$ as observation model. 

\subsection{Shape Encoding}
Our shape encoding needs to be independent of the object position and 
contain the necessary information for predicting the effect of pushes, i.e.\ the possible contact points and 
the surface normals. We use the object-centric top-down projection of the segmentation mask and depth values
to compute the $x$ and $y$ coordinates of each object pixel in this frame.
Together with the mask, the coordinates are the input to a CNN that predicts the object outline and
the unit 2D surface normals to each point on the outline. Figure \ref{fig:overview} shows
an example of the resulting $100\times100\times5$ image (outline, coordinates and normals) 
under Shape Encoding.

\subsection{Affordance Prediction}\label{sec:methods-affordance}
For each point on the object outline, the affordance module makes an approximate prediction
of the object motions that can be achieved by pushing there. For this, it densely evaluates a 
predictive model for a fixed set of representative pushing motions. This prediction then informs 
contact point selection for pushing the object towards the target. 

For our experiments, we use a relatively large set of ten representative pushes: we take five directions 
relative to the respective surface normal ($0^{\circ}$, $\pm 30^{\circ}$ and 
$\pm 60^{\circ}$) with two lengths each (1\,cm and 5\,cm). 
In general, the expressiveness of the affordance model is a tuning parameter of our method
that trades off accuracy against computational speed.\footnote{Ablation studies 
(not included for space) showed that including fewer pushes has an overall small effect. 
Removing the 5\,cm pushes was worst, increasing the average number of steps taken by more than 15\,\%.}

We evaluate two predictive models for obtaining the affordances,
the analytical model and a learned model.

\subsubsection{Affordances from the Analytical Model}
Given the representative pushes, the shape encoding and parameters $\mathbf{c}$, $l$ and $\mu$ from 
the state estimation module, we can apply the analytical model~\cite{model} at each contact point. 
We use a one-step prediction, which can be done efficiently on GPU but is less accurate than rolling 
out the model over smaller substeps:
During the push, values like the contact point and normal there can change, e.g.\ when the 
the pusher slides along the object or even loses contact completely. Such changes of the model's input values
cannot be taken into account without substeps.

\subsubsection{Affordances from a Learned Model}
Alternatively, we train a CNN to predict object movement given the pushes, $\mathbf{c}$ and the shape encoding. 
Different from the analytical model, it does not require the parameters $l$ and $\mu$, but can take the 
shape around the contact point into account to predict effects of pusher sliding like loss of contact.
For this, the model uses a 3-layer CNN with max-pooling to process the object outline. 
The resulting local shape features, the pushes and the shape encoding serve as input for 
predicting the object motion using a second 3-layer CNN without pooling. 

\subsection{Planning}
We use a greedy planner to find the contact point and straight pushing motion 
that brings the object closest to the desired goal pose at each step. We found this approach 
to be sufficient in our scenario where no obstacles are present. 
For planning around obstacles, our model could be combined with a global planner 
for object poses, e.g.\ \cite{multistep,in-hand}.
  
Instead of jointly optimizing contact point and pushing motion, we divide the problem into
two subtasks. We first propose a set of contact points and then separately optimize the pushing motions
at each candidate point before selecting the most promising combination.

\subsubsection{Contact Point Proposal}
Our method uses the affordances to score each point on the object outline by how 
close pushing there could bring the object to the target pose:
\begin{equation}\label{eq:1}
s(\mathbf{r}_i) = \min_{\mathbf{u} \in U_i} \parallel \mathbf{v}_d - \mathbf{v}_p(\mathbf{u}, r_i) \parallel_2 
+ \lambda | \dot{\theta}_d - \dot{\theta}_p(\mathbf{u}, r_i)| 
\end{equation}
Here, $\mathbf{v}_d$ and $\dot{\theta}_d$ are the desired object translation and rotation, 
$U_i$ are the representative pushing motions at contact point $\mathbf{r}_i$, and 
($\mathbf{v}_p(\mathbf{u}, r_i)$, $\dot{\theta}_p(\mathbf{u}, r_i)$) 
their predicted object motion.
We weight the rotation error (in degree) stronger ($\lambda = 2$) for a good trade-off between translation and rotation. 
A softmax function turns the scores, $s(\mathbf{r}_i)$, into a probability distribution that is used
to sample $k$ candidate points.
We found that sampling the contact points instead of choosing the $k$ best points improved the robustness
of our method.

\subsubsection{Push Motion Optimization}\label{sec:methods-optimization}

The discrete set of actions evaluated for the affordance model will in general not contain the optimal 
pushing motion at each point. We thus optimize push direction and length
at each candidate contact point by interpolating between five base pushes $U_b$ with different 
directions as follows:

\begin{enumerate}[leftmargin=*]
\item For each $\mathbf{u_b} \in U_b$, roll out the analytical model over the
max.\ push length of 5\,cm in substeps of 0.5\,cm. 
\item At each step, score the predicted object movement so far using Equation~\ref{eq:1}. 
\item Truncate each $\mathbf{u_b}$ at its best-scoring step (min. push length 1\,cm). This gives $\mathbf{\bar{u}_b}$ with optimal scores $\bar{s}_b$. 
\item Find the optimal push direction $\mathbf{u_d}$ by interpolating between the $\mathbf{\bar{u}_b}$ with
the best $\bar{s}_b$ and the two $\mathbf{\bar{u}_b}$ with neighbouring directions.
\item Optimize the length of $\mathbf{u_d}$ as in steps (1) - (3) to
find the optimal push $\mathbf{u^*}$ with score $s^*$
\end{enumerate}

As explained before, rolling out the analytical model over shorter substeps is more accurate than predicting 
the outcome of the full 5\,cm push in one step.
The planner finally returns the contact point and action with the highest $s^*$.

In our specific case, the affordances already contain predictions for the same 
five push directions that we also use for $U_b$. 
This allows us to use the affordance predictions in step (1) of the 
push optimization. Instead of steps (2,3) and (5), the push length is then
optimized by linearly rescaling the push and prediction to match the desired motion.
We compare this to our regular method in Experiment \ref{exp3}.

\begin{figure}[tbp]
\centering
\includegraphics[width=\linewidth]{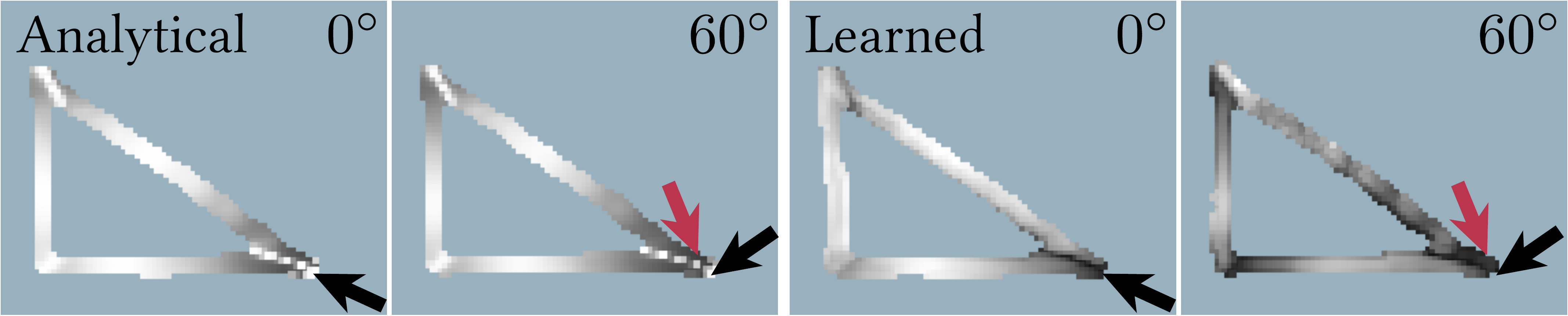}
\caption{\label{fig:affordance}
Predicted translation magnitude from the analytical and learned affordance model (brighter is higher) for
pushes along the normal and at a 60$^{\circ}$ angle. In contrast to the analytical model, the learned model
predicts low magnitude for pushes that are unlikely to properly hit the object (black arrows) or pushes that 
will slide off the object (red arrows). 
}
\end{figure}

\begin{figure}[tbp]
\centering
\includegraphics[height=1.7cm]{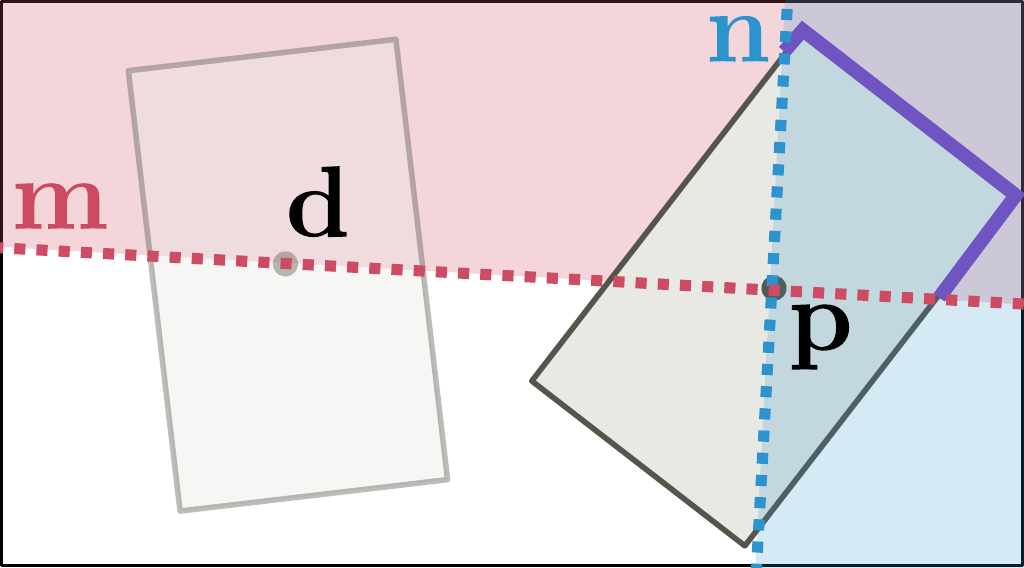}
\caption{\label{fig:baseline}
Heuristic for contact point selection (\textit{geo}): line $\mathbf{m}$
connects the current $\mathbf{p}$ and desired object position $\mathbf{d}$, $\mathbf{n}$ is its normal. 
Points on the blue side of $\mathbf{n}$ afford pushing towards $\mathbf{d}$, points to the right of $\mathbf{m}$ (red area) 
are proposed for counter-clockwise rotation. For rotations below $2^{\circ}$, candidates need to lie 
within 2\,cm of $\mathbf{m}$. The intersection of both areas (purple) defines the set of
possible contact points for sampling.} 
\end{figure}

\section{Training}

For training the perception, shape encoding and learned affordance model, we rely mostly 
on simulated data generated in pybullet\,\cite{pybullet}. 
Each datapoint contains an RGBD image of an object on a surface, its ground truth position, segmentation mask and
outline with normals. We annotate 20 random contact points per object with the object movement in 
response to the ten representative pushing actions defined in Section \ref{sec:methods-affordance}. 
Properties like object mass, centre of mass and friction coefficients are sampled randomly. 
We generate more than 15k examples using 21 objects, of which we hold out three for testing (shown
in Figure \ref{fig:objects}, which also shows the real-world setup after which we modeled the simulation).
While the segmentation and shape encoding network are finetuned on real data from the Omnipush 
dataset\,\cite{omnipush}, the learned affordance model is only trained on simulated data. 
We train the models in tensorflow\,\cite{tensorflow} using Adam\,\cite{adam}.

\section{Simulation Experiments}

\subsection{Setup}
We evaluate three different tasks: translating the object by 20\,cm without changing the orientation 
(\textit{translation}), rotating the object by 0.5\,rad ($28.6^{\circ}$) without changing the position 
(\textit{rotation}) and translating for 10\,cm plus rotating by 0.35\,rad ($20^{\circ}$) (\textit{mixed}). 
A trial counts as successful if it brings the object within less than 0.75\,cm of the desired position and
$5^{\circ}$ of the desired orientation in at most 30 steps. We evaluate the percentage of successful trials 
and the average number of steps until the goal pose is reached. 

For each task, object and method, we perform 60 trials. At the beginning of each, the object is 
placed at the center of the workspace. We vary its initial orientation in 20 steps from 0 to 360$^\circ$ and perform
three runs with each orientation.

\begin{figure*}[tbp]
\noindent
\begin{subfigure}{0.425\textwidth}
\begin{tikzpicture}
	\begin{groupplot}[
		ybar=0.pt,
		group style={group size=3 by 1,
            		  horizontal sep=0.11cm,
            	      vertical sep=0.25cm,
            		  x descriptions at=edge bottom},
        enlarge x limits=0.175,
        enlarge y limits=0,
        width=4.cm,
        height=3.7cm,
        tick label style={/pgf/number format/fixed, font=\tiny}, 
        log ticks with fixed point,
        xtick={data},
        yticklabels={0, 0, $\frac{1}{2}$, 1},
        ylabel style={yshift=-0.875cm, font=\scriptsize, inner sep=0pt, outer sep=0pt},
        yticklabel style={xshift=0.075cm},
        xlabel style={yshift=0.25cm, font=\scriptsize},
        xlabel={contact points}, 
        xticklabel style={yshift=0.075cm},
        ymax=1.1, ymin=0,
        title style={yshift=-0.15cm, font=\scriptsize},
        symbolic x coords={1, 3, 5, 10}
        ]
		
	\nextgroupplot[ylabel style={align=center}, ylabel={Success Rate \%}, title={Mixed}, bar width=0.125cm]
	  \addplot+[mpi-grey!60!black, fill=mpi-grey!95] table [x=cp, y=avg_rdn] {figures/mixed_suc2}; 
      \addplot+[mpi-grey!40!black, fill=mpi-grey!65!black] table [x=cp, y=avg_geo] {figures/mixed_suc2}; 
      \addplot+[mpi-red!70!black, fill=mpi-red!70] table [x=cp, y=avg_ana] {figures/mixed_suc2}; 
      \addplot+[mpi-red!50!black, fill=mpi-red!80!black] table [x=cp, y=avg_lrd] {figures/mixed_suc2}; 
      
    \nextgroupplot[ylabel style={align=center}, title={Translation}, yticklabels=\empty,
    				legend style={draw=none, % get rid of the box
    							   inner sep=0pt, outer sep=0pt,
								   every node/.style={inner sep=0pt, outer sep=0pt},
        			             /tikz/every even column/.append style={column sep=0.75em},  % increase spacing between legend entries (but not between picture and name)
        			   			  font=\scriptsize, at={(0.5,-0.35)}, anchor=north},
					legend columns=-1,
					legend entries={\textit{rdn}, \textit{geo}, \textit{ana}, \textit{lrn}},
					bar width=0.125cm] 
      \addplot+[mpi-grey!60!black, fill=mpi-grey!95] table [x=cp, y=avg_rdn] {figures/tr_suc2}; 
      \addplot+[mpi-grey!40!black, fill=mpi-grey!65!black] table [x=cp, y=avg_geo] {figures/tr_suc2}; 
      \addplot+[mpi-red!70!black, fill=mpi-red!70] table [x=cp, y=avg_ana] {figures/tr_suc2}; 
      \addplot+[mpi-red!50!black, fill=mpi-red!80!black] table [x=cp, y=avg_lrd] {figures/tr_suc2}; 
        
  \nextgroupplot[ylabel style={align=center}, title={Rotation}, yticklabels=\empty, bar width=0.125cm]
      \addplot+[mpi-grey!60!black, fill=mpi-grey!95] table [x=cp, y=avg_rdn] {figures/rot_suc2}; 
      \addplot+[mpi-grey!40!black, fill=mpi-grey!65!black] table [x=cp, y=avg_geo] {figures/rot_suc2}; 
      \addplot+[mpi-red!70!black, fill=mpi-red!70] table [x=cp, y=avg_ana] {figures/rot_suc2}; 
      \addplot+[mpi-red!50!black, fill=mpi-red!80!black] table [x=cp, y=avg_lrd] {figures/rot_suc2}; 
    \end{groupplot}
\end{tikzpicture}
\end{subfigure} \hspace*{0.335cm}
\begin{subfigure}{0.55\textwidth}
\begin{tikzpicture}
	\begin{groupplot}[
	    ybar=0.pt,
		group style={group size=3 by 1,
            		  horizontal sep=0.11cm,
            	      vertical sep=0.25cm,
            		  x descriptions at=edge bottom},
        enlarge x limits=0.175,
        enlarge y limits=0,
     	 width=4.55cm,
     	 height=3.7cm, 
        tick label style={/pgf/number format/fixed, font=\tiny}, 
        yticklabel style={xshift=0.075cm},
        log ticks with fixed point,
        xtick={data},
        ylabel style={yshift=-0.8cm, font=\scriptsize, inner sep=0pt, outer sep=0pt},
        xlabel style={yshift=0.25cm, font=\scriptsize},
        xlabel={contact points}, 
        xticklabel style={yshift=0.075cm},
        ymax=25, ymin=0,
        title style={yshift=-0.15cm, font=\scriptsize},
        symbolic x coords={1, 3, 5, 10}
		]
		
	\nextgroupplot[ylabel style={align=center}, ylabel={Steps}, title={Mixed}, bar width=0.15cm]
	  \addplot+[mpi-grey!60!black, fill=mpi-grey!95,
				 error bars/.cd, y dir=both, y explicit, error bar style={line width=0.9pt}] table
        [x=cp, y=avg_rdn, y error=avg_rdn_std] {figures/mixed_steps_suc2}; 
      \addplot+[mpi-grey!40!black, fill=mpi-grey!65!black,
      			 error bars/.cd, y dir=both, y explicit, error bar style={line width=0.9pt}] table
        [x=cp, y=avg_geo, y error=avg_geo_std] {figures/mixed_steps_suc2}; 
      \addplot+[mpi-red!70!black, fill=mpi-red!70,
      			 error bars/.cd, y dir=both, y explicit, error bar style={line width=0.9pt}] table
        [x=cp, y=avg_ana, y error=avg_ana_std] {figures/mixed_steps_suc2}; 
      \addplot+[mpi-red!50!black, fill=mpi-red!80!black,
      			 error bars/.cd, y dir=both, y explicit, error bar style={line width=0.9pt}] table
        [x=cp, y=avg_lrd, y error=avg_lrd_std] {figures/mixed_steps_suc2};

	\nextgroupplot[ylabel style={align=center}, title={Translation}, yticklabels=\empty,
					legend style={draw=none, % get rid of the box
									inner sep=0pt, outer sep=0pt,
								   every node/.style={inner sep=0pt, outer sep=0pt},
        			             /tikz/every even column/.append style={column sep=0.75em},  % increase spacing between legend entries (but not between picture and name)
        			   			  font=\scriptsize, at={(0.5,-0.35)}, anchor=north},
			        legend columns=-1,
					legend entries={\textit{rdn}, \textit{geo}, \textit{ana}, \textit{lrn}},
					 bar width=0.15cm]
	  \addplot+[mpi-grey!60!black, fill=mpi-grey!95, error bars/.cd, y dir=both, y explicit, error bar style={line width=0.9pt}] table
        [x=cp, y=avg_rdn, y error=avg_rdn_std] {figures/tr_steps_suc2}; 
      \addplot+[mpi-grey!40!black, fill=mpi-grey!65!black, error bars/.cd, y dir=both, y explicit, error bar style={line width=0.9pt}] table
        [x=cp, y=avg_geo, y error=avg_geo_std] {figures/tr_steps_suc2}; 
      \addplot+[mpi-red!70!black, fill=mpi-red!70, error bars/.cd, y dir=both, y explicit, error bar style={line width=0.9pt}] table
        [x=cp, y=avg_ana, y error=avg_ana_std] {figures/tr_steps_suc2}; 
      \addplot+[mpi-red!50!black, fill=mpi-red!80!black, error bars/.cd, y dir=both, y explicit, error bar style={line width=0.9pt}] table
        [x=cp, y=avg_lrd, y error=avg_lrd_std] {figures/tr_steps_suc2}; 
        
	\nextgroupplot[ylabel style={align=center}, title={Rotation}, yticklabels=\empty, bar width=0.15cm]
	  \addplot+[mpi-grey!60!black, fill=mpi-grey!95, error bars/.cd, y dir=both, y explicit, error bar style={line width=0.9pt}] table
        [x=cp, y=avg_rdn, y error=avg_rdn_std] {figures/rot_steps_suc2}; 
      \addplot+[mpi-grey!40!black, fill=mpi-grey!65!black, error bars/.cd, y dir=both, y explicit, error bar style={line width=0.9pt}] table
        [x=cp, y=avg_geo, y error=avg_geo_std] {figures/rot_steps_suc2}; 
      \addplot+[mpi-red!70!black, fill=mpi-red!70, error bars/.cd, y dir=both, y explicit, error bar style={line width=0.9pt}] table
        [x=cp, y=avg_ana, y error=avg_ana_std] {figures/rot_steps_suc2}; 
      \addplot+[mpi-red!50!black, fill=mpi-red!80!black, error bars/.cd, y dir=both, y explicit, error bar style={line width=0.9pt}] table
        [x=cp, y=avg_lrd, y error=avg_lrd_std] {figures/rot_steps_suc2}; 
        
    \end{groupplot}
\end{tikzpicture}
\end{subfigure}
\caption{\label{fig:cp}Pushing performance over number of sampled contact points. We compare different sampling 
methods of contact locations: randomly (\textit{rdn}) or according to the geometric baseline (\textit{geo}), analytical (\textit{ana}) 
or learned (\textit{lrn}) affordances. We analyse performance for three different tasks: Pure object translation, 
pure object rotation and a mixed motion. Results are averaged over three test objects (See Fig.~\ref{fig:objects}). 
Our proposed affordance model (either \textit{ana} or \textit{lrn}) generally requires only one contact point sample 
to achieve a high success rate with the lowest number of steps to get to a target object pose.}
\end{figure*}
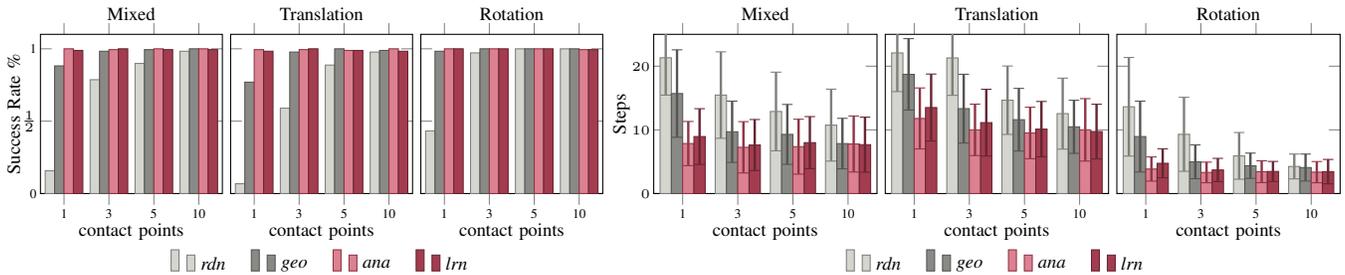

\subsection{Affordance Prediction}\label{exp1}
We first qualitatively compare the learned and the analytical affordance model to see if there are any major 
differences between them.
Overall, both models predict similar directions of movement, with the analytical model predicting more
pronounced rotation. A potential advantage of using a learned model becomes apparent when we compare the magnitude
of the predicted translational movement, which is shown in Figure \ref{fig:affordance}.
The analytical model predicts strong translation for pushes to the sharp corners of the triangle, whereas the 
learned model predicts comparatively low magnitudes there. The same effect is visible for angled pushes that cause
the pusher to slide towards corners. 
As discussed before, the analytical affordance model cannot predict a loss of contact due to pusher sliding. 
However, this is more likely when pushing at sharp corners or with high angles. The learned model takes the object 
shape around the contact point into account and is therefore better at predicting such cases.

\subsection{Contact Point Selection}\label{exp2}
In this experiment, we test our hypothesis that explicitly reasoning about the contact locations
makes planning more efficient as compared to sampling actions that collide with the object in random locations. 
For this, we vary the number of sampled contact points and compare our approach (that uses the affordances 
to propose promising contact points) to two baselines that select the contact points more randomly.

The simplest baseline samples uniformly from all points on the object outline (\textit{rdn}). 
A more informed approach (\textit{geo}) uses a geometric heuristic explained in Figure \ref{fig:baseline}.  
Based on the desired motion, it defines a quadrant of the object from which the contact points are sampled.
In contrast to \textit{rdn}, \textit{geo} better avoids sampling contact points at which the object can only be pushed 
away from the goal. It however ignores the exact shape of the object and can thus still propose unsuitable 
contact locations especially for non-convex objects.

To minimize the influence of other components of our system on the results,  we do not use filtering
for state estimation in this experiment but assume access to perfect state information at every step.

\subsubsection*{Results}
We first compare the success rates in Figure \ref{fig:cp} (left). By sampling from the 
affordance model (learned \textit{lrn} or analytical \textit{ana}), our method can already achieve a success rate close 
to 100\% with only one contact point. The geometric heuristic also performs well and often reaches 100\% with as 
few as three contact points. We only see a big impact of the number of contact points when sampling randomly. On the tasks that involve translation, 
\textit{rdn} only reaches the success rate of the other methods with ten contact points. The number of sampled 
points is generally more important for translating than for rotating.

Figure \ref{fig:cp} (right) shows the number of steps each method took until the goal pose was reached. Even in 
successful runs, \textit{rdn} needs significantly more steps than the other methods. \textit{Geo} again performs better, 
although still worse than our proposed method using the affordance prediction.
Both \textit{lrn} and \textit{ana} work very well with only one contact point
and their performance mostly saturates at three sampled candidates. There is no significant
difference between using the analytical or the learned model for obtaining the affordances. 

To summarize, using an affordance model to sample contact points makes planning more efficient by reducing
the number of contact points that have to be evaluated per step and the number of steps taken until the goal is 
reached.

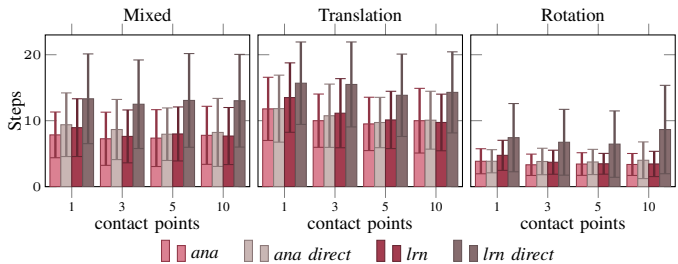
\begin{figure}[htbp]
\hspace*{-0.375cm}
\noindent
\begin{tikzpicture}
	\begin{groupplot}[
		group style={
            % place the four plots next to each other
            group size=3 by 1,
            horizontal sep=0.11cm,
            vertical sep=0.25cm,
            x descriptions at=edge bottom,
            },
        enlarge x limits=0.175,
        enlarge y limits=0,
        width=4.3cm,
        height=3.6cm,
        tick label style={/pgf/number format/fixed, font=\tiny}, 
        log ticks with fixed point,
        xtick={data},
        ylabel style={yshift=-0.85cm, font=\scriptsize, inner sep=0pt, outer sep=0pt},
        xlabel style={yshift=0.25cm, font=\scriptsize},
        xlabel={contact points},  
        ymax=23, ymin=0,
        title style={yshift=-0.15cm, font=\scriptsize},
        ybar=0pt,
        symbolic x coords={1, 3, 5, 10},
        xticklabel style={yshift=0.075cm},
        yticklabel style={xshift=0.075cm},
        ]
        
	\nextgroupplot[ylabel style={align=center}, ylabel={Steps}, title={Mixed}, bar width=0.145cm]
	  \addplot+[mpi-red!70!black, fill=mpi-red!70, error bars/.cd, y dir=both, y explicit, error bar style={line width=0.9pt}] table
        [x=cp, y=avg_ana, y error=avg_ana_std] {figures/mixed_steps_suc2}; 
      \addplot[mpi-grey!60!black!85!mpi-red, fill=mpi-grey!95!black!85!mpi-red, error bars/.cd, y dir=both, y explicit, error bar style={line width=0.9pt}] table
        [x=cp, y=avg_no_mpc_ana, y error=avg_no_mpc_ana_std] {figures/mixed_steps_suc2}; 
      \addplot[mpi-red!50!black, fill=mpi-red!80!black, error bars/.cd, y dir=both, y explicit, error bar style={line width=0.9pt}] table
        [x=cp, y=avg_lrd, y error=avg_lrd_std] {figures/mixed_steps_suc2}; 
      \addplot[mpi-grey!40!black!85!mpi-red, fill=mpi-grey!55!black!80!mpi-red, error bars/.cd, y dir=both, y explicit, error bar style={line width=0.9pt}] table
        [x=cp, y=avg_no_mpc_lrd, y error=avg_no_mpc_lrd_std] {figures/mixed_steps_suc2};

    \nextgroupplot[ylabel style={align=center}, title={Translation}, yticklabels=\empty,
    				legend style={draw=none, % get rid of the box
    								inner sep=0pt, outer sep=0pt,
								   every node/.style={inner sep=0pt, outer sep=0pt},
        			             /tikz/every even column/.append style={column sep=0.75em},  % increase spacing between legend entries (but not between picture and name)
        			   			  font=\scriptsize, at={(0.5,-0.35)}, anchor=north},
			        legend columns=-1,
					 legend entries={\textit{ana}, \textit{ana direct}, \textit{lrn}, \textit{lrn direct}},
					 bar width=0.145cm]
	  \addplot+[mpi-red!70!black, fill=mpi-red!70, error bars/.cd, y dir=both, y explicit, error bar style={line width=0.9pt}] table
        [x=cp, y=avg_ana, y error=avg_ana_std] {figures/tr_steps_suc2};
      \addplot[mpi-grey!60!black!85!mpi-red, fill=mpi-grey!95!black!85!mpi-red, error bars/.cd, y dir=both, y explicit, error bar style={line width=0.9pt}] table
        [x=cp, y=avg_no_mpc_ana, y error=avg_no_mpc_ana_std] {figures/tr_steps_suc2};  
      \addplot[mpi-red!50!black, fill=mpi-red!80!black, error bars/.cd, y dir=both, y explicit, error bar style={line width=0.9pt}] table
        [x=cp, y=avg_lrd, y error=avg_lrd_std] {figures/tr_steps_suc2}; 
      \addplot[mpi-grey!40!black!85!mpi-red, fill=mpi-grey!55!black!80!mpi-red, error bars/.cd, y dir=both, y explicit, error bar style={line width=0.9pt}] table
        [x=cp, y=avg_no_mpc_lrd, y error=avg_no_mpc_lrd_std] {figures/tr_steps_suc2};     
        
  \nextgroupplot[ylabel style={align=center}, title={Rotation}, yticklabels=\empty, bar width=0.145cm]
      \addplot+[mpi-red!70!black, fill=mpi-red!70, error bars/.cd, y dir=both, y explicit, error bar style={line width=0.9pt}] table
        [x=cp, y=avg_ana, y error=avg_ana_std] {figures/rot_steps_suc2}; 
      \addplot[mpi-grey!60!black!85!mpi-red, fill=mpi-grey!95!black!85!mpi-red, error bars/.cd, y dir=both, y explicit, error bar style={line width=0.9pt}] table
        [x=cp, y=avg_no_mpc_ana, y error=avg_no_mpc_ana_std] {figures/rot_steps_suc2}; 
      \addplot[mpi-red!50!black, fill=mpi-red!80!black, error bars/.cd, y dir=both, y explicit, error bar style={line width=0.9pt}] table
        [x=cp, y=avg_lrd, y error=avg_lrd_std] {figures/rot_steps_suc2}; 
      \addplot[mpi-grey!40!black!85!mpi-red, fill=mpi-grey!55!black!80!mpi-red, error bars/.cd, y dir=both, y explicit, error bar style={line width=0.9pt}] table
        [x=cp, y=avg_no_mpc_lrd, y error=avg_no_mpc_lrd_std] {figures/rot_steps_suc2};   
 
    \end{groupplot}
\end{tikzpicture}
\caption{\label{fig:no_mpc}Steps taken vs. sampled contact points when rolling out the analytical model for optimizing the
push motions (\textit{ana, lrn}) or directly using the affordance (\textit{ana-direct, lrn-direct}). While \textit{ana, lrn} 
and \textit{ana-direct} perform similar, \textit{lrn-direct} is less accurate and thus needs more steps to succeed.}
\end{figure}

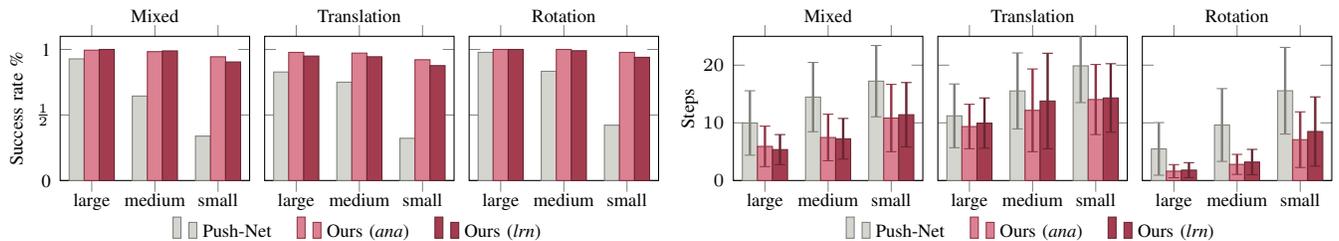
\begin{figure*}[htbp]
\begin{subfigure}{0.46\textwidth}
\begin{tikzpicture}
	\begin{groupplot}[
		group style={
            % place the plots next to each other
            group size=3 by 1,
            horizontal sep=0.2cm,
            x descriptions at=edge bottom,
            },
        enlarge x limits=0.25,
        enlarge y limits=0,
        width=4.1cm,
        height=3.5cm,
        tick label style={/pgf/number format/fixed, font=\scriptsize}, 
        log ticks with fixed point,
        xtick={data},
        ylabel style={yshift=-0.65cm, font=\scriptsize},
        yticklabels={0, 0, $\frac{1}{2}$, 1},
        xticklabel style={yshift=0.1cm},
        ymax=1.1, ymin=0,
        ybar=0.pt,
        symbolic x coords={large, medium, small},
        title style={yshift=-0.15cm, font=\scriptsize}]
		
	\nextgroupplot[ylabel style={align=center}, title={Mixed}, ylabel={Success rate \%}, bar width=0.2cm]
		\addplot[mpi-grey!60!black, fill=mpi-grey!95, fill=mpi-grey] table [x=cp, y=avg_pn_suc] {figures/com_mixed_suc}; 
      \addplot[mpi-red!70!black, fill=mpi-red!70] table [x=cp, y=avg_ana_suc] {figures/com_mixed_suc}; 
      \addplot[mpi-red!50!black, fill=mpi-red!80!black] table [x=cp, y=avg_lrn_suc] {figures/com_mixed_suc}; 
        
    \nextgroupplot[ylabel style={align=center}, title={Translation}, yticklabels=\empty,
    				legend style={draw=none, % get rid of the box
    							  inner sep=0pt, outer sep=0pt,
								  every node/.style={inner sep=0pt, outer sep=0pt},
        			             /tikz/every even column/.append style={column sep=0.75em},  % increase spacing between legend entries (but not between picture and name)
        			   			  font=\scriptsize, at={(0.5,-0.25)}, anchor=north},
			        legend columns=-1,
					 legend entries={Push-Net, Ours (\textit{ana}), Ours (\textit{lrn})}, 
					bar width=0.2cm]
      \addplot[mpi-grey!60!black, fill=mpi-grey!95] table [x=cp, y=avg_pn_suc] {figures/com_tr_suc}; 
      \addplot[mpi-red!70!black, fill=mpi-red!70] table [x=cp, y=avg_ana_suc] {figures/com_tr_suc}; 
      \addplot[mpi-red!50!black, fill=mpi-red!80!black] table [x=cp, y=avg_lrn_suc] {figures/com_tr_suc}; 
        
  \nextgroupplot[ylabel style={align=center}, title={Rotation}, yticklabels=\empty, bar width=0.2cm]
      \addplot[mpi-grey!60!black, fill=mpi-grey!95] table [x=cp, y=avg_pn_suc] {figures/com_rot_suc}; 
      \addplot[mpi-red!70!black, fill=mpi-red!70] table [x=cp, y=avg_ana_suc] {figures/com_rot_suc}; 
      \addplot[mpi-red!50!black, fill=mpi-red!80!black] table [x=cp, y=avg_lrn_suc] {figures/com_rot_suc}; 
    \end{groupplot}
\end{tikzpicture}
\end{subfigure} \hspace*{0.5cm}
\begin{subfigure}{0.46\textwidth}
%\centering
\begin{tikzpicture}
	\begin{groupplot}[
		group style={
            % place the plots next to each other
            group size=3 by 1,
            horizontal sep=0.2cm,
            x descriptions at=edge bottom,
            },
        enlarge x limits=0.25,
        enlarge y limits=0,
     	 width=4.1cm,
     	 height=3.5cm, 
        tick label style={/pgf/number format/fixed, font=\scriptsize}, 
        log ticks with fixed point,
        xtick={data},
        ylabel style={yshift=-0.65cm, font=\scriptsize},
        xticklabel style={yshift=0.1cm},
        ybar=0.pt,
        ymax=25, ymin=0,
        symbolic x coords={large, medium, small},
        title style={yshift=-0.15cm, font=\scriptsize}
		]
		
	\nextgroupplot[ylabel style={align=center}, title={Mixed}, ylabel={Steps}, bar width=0.2cm]
	  \addplot[mpi-grey!60!black, fill=mpi-grey!95, error bars/.cd, y dir=both, y explicit, error bar style={line width=0.9pt}] table
        [x=cp, y=avg_pn, y error=avg_pn_std] {figures/com_mixed_suc}; 
      \addplot[mpi-red!70!black, fill=mpi-red!70, error bars/.cd, y dir=both, y explicit, error bar style={line width=0.9pt}] table
        [x=cp, y=avg_ana, y error=avg_ana_std] {figures/com_mixed_suc}; 
      \addplot[mpi-red!50!black, fill=mpi-red!80!black, error bars/.cd, y dir=both, y explicit, error bar style={line width=0.9pt}] table
        [x=cp, y=avg_lrn, y error=avg_lrn_std] {figures/com_mixed_suc}; 
        
	\nextgroupplot[ylabel style={align=center}, title={Translation}, yticklabels=\empty,
					legend style={draw=none, % get rid of the box
								   inner sep=0pt, outer sep=0pt,
								   every node/.style={inner sep=0pt, outer sep=0pt},
        			             /tikz/every even column/.append style={column sep=0.75em},  % increase spacing between legend entries (but not between picture and name)
        			   			  font=\scriptsize, at={(0.5,-0.25)}, anchor=north},
			        legend columns=-1,
					 legend entries={Push-Net, Ours (\textit{ana}), Ours (\textit{lrn})},
					 bar width=0.2cm]
	  \addplot[mpi-grey!60!black, fill=mpi-grey!95, error bars/.cd, y dir=both, y explicit, error bar style={line width=0.9pt}] table
        [x=cp, y=avg_pn, y error=avg_pn_std] {figures/com_tr_suc}; 
      \addplot[mpi-red!70!black, fill=mpi-red!70, error bars/.cd, y dir=both, y explicit, error bar style={line width=0.9pt}] table
        [x=cp, y=avg_ana, y error=avg_ana_std] {figures/com_tr_suc}; 
      \addplot[mpi-red!50!black, fill=mpi-red!80!black, error bars/.cd, y dir=both, y explicit, error bar style={line width=0.9pt}] table
        [x=cp, y=avg_lrn, y error=avg_lrn_std] {figures/com_tr_suc}; 
        
	\nextgroupplot[ylabel style={align=center}, title={Rotation}, yticklabels=\empty, bar width=0.2cm]
	  \addplot[mpi-grey!60!black, fill=mpi-grey!95, error bars/.cd, y dir=both, y explicit, error bar style={line width=0.9pt}] table
        [x=cp, y=avg_pn, y error=avg_pn_std] {figures/com_rot_suc}; 
      \addplot[mpi-red!70!black, fill=mpi-red!70, error bars/.cd, y dir=both, y explicit, error bar style={line width=0.9pt}] table
        [x=cp, y=avg_ana, y error=avg_ana_std] {figures/com_rot_suc}; 
      \addplot[mpi-red!50!black, fill=mpi-red!80!black, error bars/.cd, y dir=both, y explicit, error bar style={line width=0.9pt}] table
        [x=cp, y=avg_lrn, y error=avg_lrn_std] {figures/com_rot_suc}; 
    \end{groupplot}
\end{tikzpicture}
\end{subfigure}
\caption{\label{fig:com} Performance of our method and Push-Net on objects with random COM (averaged over objects). We evaluate
three goal region sizes, \textit{small} (0.75\,cm 5$^{\circ}$), \textit{medium} (2.5\,cm 7.5$^{\circ}$) and \textit{large} 
(5\,cm 10$^{\circ}$). Our method has a higher success rate on smaller goal regions and needs fewer steps to reach the goal.}
\end{figure*}

\subsection{Pushing Motion Optimization}\label{exp3}

In this experiment, we test if the predicted affordances are accurate
enough to also be used for optimizing the pushing actions (see Section \ref{sec:methods-optimization}). This is especially 
interesting for evaluating the quality of the learned model.
We call the variants that use the predictions from the affordance model directly \textit{ana direct} 
and \textit{lrn direct} respectively. 

\subsubsection*{Results}

As shown in Figure \ref{fig:no_mpc}, using the analytical affordance predictions does not significantly increase the 
number of steps as compared to \textit{ana} which optimizes the actions by rolling out the analytical model over 
smaller substeps. When using the learned affordances for push optimization, the number of steps increases 
by up to four and the success rate drops by up to 10\% compared to \textit{lrn}. 
This implies that while being sufficient for selecting contact points, 
the learned model is not as accurate as the analytical model for predicting the outcome of a push. 
The negative effect of using the learned model is also not compensated by evaluating more contact points, 
which emphasizes the value of an accurate predictive model for optimizing the pushes.

\subsection{Full System}\label{exp4}
Now we evaluate the accuracy of our full system including the state estimation module, with three
contact points sampled per step. In all experiments, $\mathbf{c}$ is initialized to zero and $l$ 
and $\mu$ to reasonable estimates. 
We first test on objects whose centre of mass coincides with the geometric center
to evaluate the perception module and how well planning works with imperfect pose information. 
In the second experiment, we verify the benefit of estimating latent properties of the object 
on the example of the COM. For this, we sample the COM uniformly inside the objects. 
We also compare to Push-Net \cite{push-net} under this condition. Push-Net uses 
top-down segmentation maps of the current and desired pose as input to evaluate randomly sampled 
actions. A local planner generates sub-goals by interpolating between 
the current and the goal pose with a fixed step size, we use 5\,cm  and 
$10^{\circ}$.

\subsubsection*{Results}

In the previous experiments, we used ground truth object pose information. Here, we compare those results to doing pose 
estimation by filtering. We find that using the filter has no impact on the success rate or the number of steps taken. 
However, it increases the (true) average end pose error from 5.0$\pm$1.8\,mm, 1.8$\pm$1.3$^{\circ}$ to 8.7$\pm$4.2\,mm and 3.0$\pm$2.2$^{\circ}$. 
This is expected as we use the \textit{estimated} object pose to determine if the goal is reached. Therefore, the real pose error 
can be higher than the (7.5\,mm, 5$^{\circ}$) margin of the goal region.

On objects with a randomly sampled COM, we first verify that estimating the
COM position is beneficial, by comparing to a variant that assumes a fixed COM. 
For the triangle and butter shape, the average estimation error for $\mathbf{c}$ is 17.7$\pm$8.5\,mm, 
on the hexagon it is around 1\,cm higher. The average distance of $\mathbf{c}$ to $\mathbf{p}$ is 37.4$\pm$12.3\,mm. 
Despite not being extremely accurate, estimating $\mathbf{c}$ increases the success rate by up 
to 5\% depending on the task and object. The number of steps is also lower, but not significantly, 
most likely because estimating the COM takes a few steps in which the object often does not move towards the goal.

We also compare our approach to Push-Net, which uses an LSTM to estimate the COM. 
We evaluate three sizes of the goal region, from the (0.75\,cm, 5$^{\circ}$) we used 
in all previous experiments to the (5\,cm, 10$^{\circ}$) used in the original Push-Net paper,
plus a medium size of (2.5\,cm, 7.5$^{\circ}$). Results are shown in Figure \ref{fig:com}.
With the largest goal region, Push-Net performs competitive to our approach and it still reaches a good success rate for the 
medium sized region. On the smallest size however, our approach outperforms Push-Net by a large margin, 
despite evaluating much fewer actions. Our method also requires less steps to reach each level of
accuracy. Qualitatively, Push-Net does well for translating the object, 
but has trouble controlling its orientation precisely. 
%\rev{relate to figures (shows more in the steps than int he success rate)}
 
\section{Real-Robot Experiments}

We also evaluate our approach on a real system (ABB IRB 120 industrial robotic arm, Intel RealSense D415 camera, see Fig.~\ref{fig:teaser}). 
This is especially interesting with respect to our predictive models: We know that the analytical model makes 
assumptions that are frequently violated in the real world, while the learned model 
was trained purely in simulation and might not transfer well to the real world. 

\subsubsection{Setup}

To evaluate our affordance models, we first compare \textit{lrn}, \textit{ana} and \textit{lrn direct} 
given ground truth state information on the butter object from the MIT Push Dataset~\cite{push_data} 
that we also used in the simulation experiments (see Fig.~\ref{fig:objects}).
Then we test the full system with the analytical affordance model on butter, triangle and a new object 
from the Omnipush Dataset~\cite{omnipush} (shown in Fig.~\ref{fig:teaser})
This objects has weights to change its pressure distribution and centre of mass and thus violates 
the uniform pressure distribution assumption of the analytical model.
For both experiments, we use a new task (12\,cm translation, 46$^{\circ}$ rotation, maximum 20 steps).
Every experiment is repeated 15 times.

\subsubsection{Results}

When comparing \textit{lrn}, \textit{ana} and \textit{lrn direct} on the
real butter object, the results are very similar to the simulation experiments.
Using the analytic push optimization step, both \textit{lrn}, \textit{ana} succeed in all trials
and need 5.5$\pm$2.6 and 5.2$\pm$2.1 steps respectively. For \textit{lrn direct}, the average 
number of steps increases to 8.2$\pm$4.3. 

With filtering, we achieve an average success rate of 97\% on triangle and butter with an 
end pose error of 9.2$\pm$4.2\,mm, 5.8$\pm2.1^{\circ}$ in 6.7$\pm$3.5 steps. For the omnipush
object, the orientation estimation sometimes fails, dropping the success rate to 89\% and increasing
the number of steps to 8.0$\pm$3.6. We still reach an average end pose error of 8.8$\pm$4.5\,mm,  
7.2$\pm5.5^{\circ}$, confirming that our approach generalizes to conditions that violate
the assumptions of the analytical model.

\FloatBarrier
\section{Conclusion}

We presented an approach for vision-based manipulation that uses
an affordance model for selecting contact points during planning.
Our experiments on planar pushing show that by explicitly reasoning over contact locations,
we evaluate less actions and plan more optimal pushing actions than
when sampling random contact locations. Our method also reaches a higher accuracy than previous 
vision-based work by relying on a physically meaningful state representation.

By comparing a learned and an analytical predictive model, we show that for 
selecting contact locations, approximate predictions are sufficient.
However, to optimize the pushing motion at each contact point, the higher accuracy of the 
analytical model proved to be important.

We thus find that using a hybrid approach - combining learned and analytical components - 
is beneficial for robotics. Learning is not only well suited for
perception but also for ``intuitive physics'' models that can quickly narrow down 
large search spaces to few promising candidates that are then optimized
using more accurate but costly analytical models. 

Limitations are the simple scenes we consider
and that our method assumes a mostly unoccluded object outline. Dealing
with strong occlusion is an interesting problem for future work.
Accurately estimating object orientation also proved challenging in some cases. Finally, 
we would like to test on more diverse, non-planar objects. 
Preliminary results in simulation suggest that our approach is robust to this, but 
real-world experiments are necessary to confirm these results.


%%%%%%%%%%%%%%%%%%%%%%%%%%%%%%%%%%%%%%%%%%%%%%%%%%%%%%%%%%%%%%%%%%%%%%%%%%%%%%%%

%%%%%%%%%%%%%%%%%%%%%%%%%%%%%%%%%%%%%%%%%%%%%%%%%%%%%%%%%%%%%%%%%%%%%%%%%%%%%%%%

%%%%%%%%%%%%%%%%%%%%%%%%%%%%%%%%%%%%%%%%%%%%%%%%%%%%%%%%%%%%%%%%%%%%%%%%%%%%%%%%
%\section*{APPENDIX}
\section*{ACKNOWLEDGMENT}
This  research  was  supported  in  part  by  the  Max-Planck-Society, the Toyota Research Institute (TRI), 
and ONR MURI N00014-16-1-2007.
Any  opinions,  findings,  and  conclusions  or recommendations expressed in this material 
are those of the authors and do not necessarily reflect the views of the funding organizations 
or any other Toyota entity. 
The authors thank the International Max Planck Research School for Intelligent Systems (IMPRS-IS) for supporting Alina Kloss.

%%%%%%%%%%%%%%%%%%%%%%%%%%%%%%%%%%%%%%%%%%%%%%%%%%%%%%%%%%%%%%%%%%%%%%%%%%%%%%%%

%{\scriptsize
\bibliographystyle{IEEEtranN}
\bibliography{paper}
%}

\end{document}